\documentclass{article}
\usepackage{spconf,amsmath,epsfig}
\usepackage{setspace}
\usepackage{color,soul}
\usepackage{cite}
\usepackage{array}
\usepackage{multirow}
\usepackage{url}
\usepackage[T1]{fontenc}

\title{MagicBathyNet: A Multimodal Remote Sensing Dataset for Bathymetry Prediction and Pixel-based Classification in Shallow Waters}
\name{Panagiotis Agrafiotis\textsuperscript{ 1, 2*}\thanks{
*Corresponding author: \texttt{agrafiotis@tu-berlin.de}}, Łukasz Janowski\textsuperscript{ 3}, Dimitrios Skarlatos\textsuperscript{ 4} and Beg{\"u}m Demir\textsuperscript{ 1, 2}}
\address{\textsuperscript{1}BIFOLD - Berlin Institute for the Foundations of Learning and Data, Berlin, Germany\\
\textsuperscript{2}Faculty of Electrical Engineering and Computer Science, Technische Universit{\"a}t Berlin, Germany\\
\textsuperscript{3}Maritime Institute, Gdynia Maritime University, Gdynia, Poland\\
\textsuperscript{4}Department of Civil Engineering and Geomatics, Cyprus University of Technology, Limassol, Cyprus}

\begin{document}
%
\maketitle
\begin{abstract}
Accurate, detailed, and regularly updated bathymetry, coupled with complex semantic content, is crucial for the under-mapped shallow water areas facing intense climatological and anthropogenic pressures. Current methods exploiting remote sensing imagery to derive bathymetry or pixel-based seabed classes mainly exploit non-open data. This lack of openly accessible benchmark archives prevents the wider use of deep learning methods in such applications. To address this issue, in this paper we present the MagicBathyNet, which is a benchmark dataset made up of image patches of Sentinel-2, SPOT-6 and aerial imagery, bathymetry in raster format and annotations of seabed classes. MagicBathyNet is then exploited to benchmark state-of-the-art methods in learning-based bathymetry and pixel-based classification. Dataset, pre-trained weights, and code are publicly available at \url{www.magicbathy.eu/magicbathynet.html}. 
\end{abstract}

\vspace{-0.05in}
\section{Introduction}
\label{sec:intro}
\vspace{-0.05in}
About 71\% of the Earth's surface is water-covered but only a small fraction is mapped by direct observation so far. The recognition of the importance of seafloor mapping is recently increased, driven by habitat destruction, pollution, natural disasters, navigation, and other needs. Accurate, detailed and high-frequent bathymetry, coupled with the complex semantic content, is crucial for the under-mapped shallow coastal areas, being affected by intense climatological and anthropogenic pressures. On the one hand, acoustic methods are inefficient in shallow waters and subject to waves, reefs, and fail due to multi-path errors, while LiDAR (Light Detection And Ranging) is expensive, both lacking the visual information \cite{agrafiotis2019, agrafiotis2020}. To overcome the above issues, aerial and satellite images are used extensively nowadays \cite{mandlburger2022review}. Spectrally Derived Bathymetry (SDB) is based on the attenuation of radiance as a function of depth and wavelength in the water and can deliver depths over large shallow areas. Additionally to the well established physics- and simple regression-based methods such as random forests, and support vector machines, deep learning-based approaches have recently gained great attention for SDB \cite{ai2020convolutional,al2021satellite,kaloop2021hybrid,lumban2022extracting,mandlburger2021bathynet,xi2023band}. Deep learning methods can handle better the complex interaction of light with the water surface, column, and water bottom compared to simple models \cite{mandlburger2022review}. On the other hand, deep learning-based pixel classification approaches of the seabed are mainly devoted to map coral reefs or seagrass meadows \cite{9984190,b2020mapping} and mainly involve fully convolutional network (FCN) architectures. The existing methods for both SDB and pixel classification are mostly applied on non-public dataset. The lack of openly accessible benchmark archives prevents the wider use of deep learning methods in such applications and limits the reliable comparison of the developed methods. To fill this gap and encourage further research endeavors in learning-based bathymetric mapping and pixel-level classification tasks, this paper introduces MagicBathyNet, which is a new multi-modal benchmark dataset. Leveraging our dataset, we benchmark state-of-the-art methods in learning-based bathymetric mapping and pixel-based classification of the seafloor.

\vspace{-0.05in}
\section{Description of MagicBathyNet}
\label{sec:descr}
\begin{figure}[h!]
    \begin{minipage}[c]{1\linewidth}
        \centering
        \includegraphics[width=7cm]{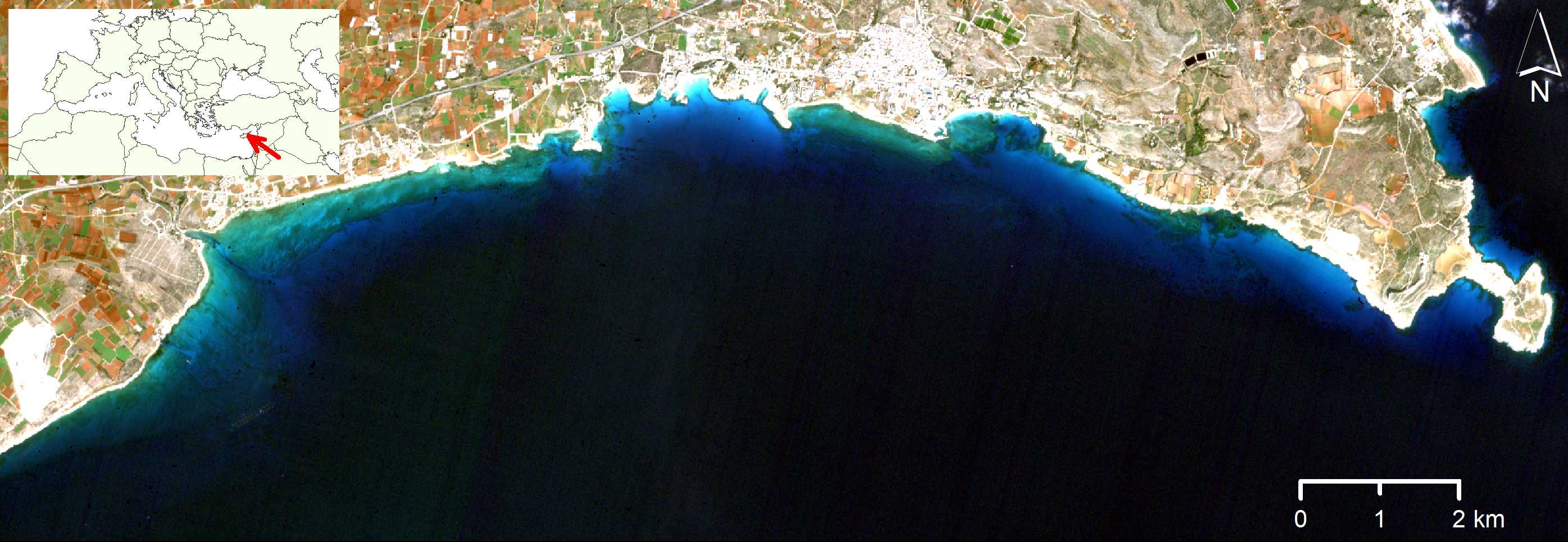}\\
        (a)
    \end{minipage}
    
    \hfill
    \begin{minipage}[c]{1\linewidth}
        \centering
        \includegraphics[width=7cm]{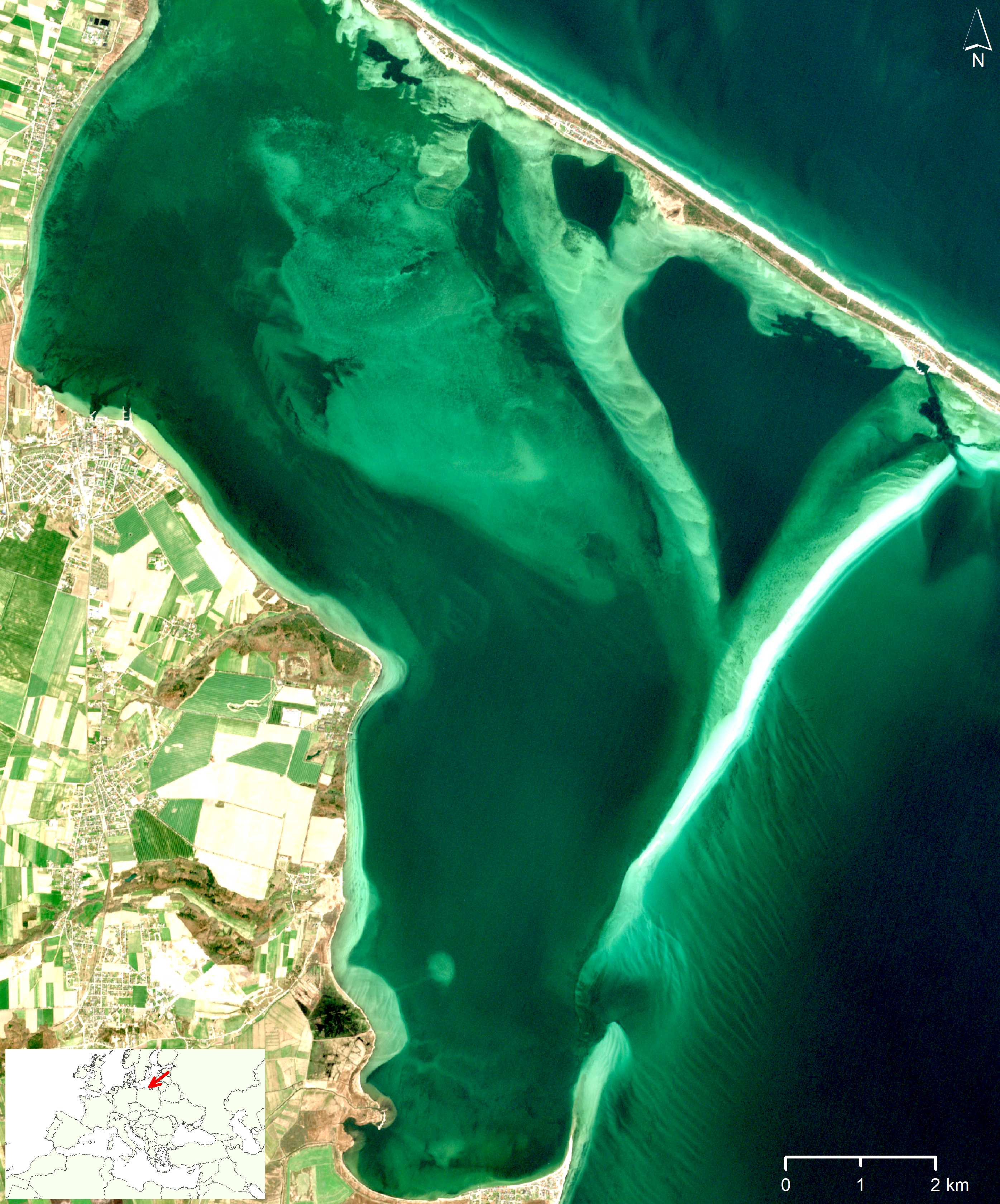}\\
        (b)
    \end{minipage}
    \hfill
    \vspace{-0.15in}
    \caption{The considered areas: (a) Agia Napa and (b) Puck Lagoon as depicted in the Sentinel-2 Level2A imagery (source: Copernicus Hub).}
    \label{fig:areas}
\end{figure}

\vspace{-0.05in}
MagicBathyNet has been designed to be geographically well-distributed. It's coverage includes two vastly different coastal areas (in terms of water column characteristics and bottom type): i) Agia Napa area in Cyprus (Fig.\ref{fig:areas}a), covering a wide range of typical Mediterranean waters and seabed types, and ii) Puck Lagoon area in Poland (Fig.\ref{fig:areas}b), representing in a great degree Baltic Sea waters and bottom. MagicBathyNet contains 3355 co-registered triplets of Sentinel-2 (S2), SPOT-6, and aerial image patches, complemented by 1244 co-registered S2 and SPOT-6 pairs, 3354 Digital Surface Model (DSM) patches for the aerial patches and 3396 DSM patches for S2 and SPOT-6. The smaller number of DSM data is a result of their coverage being limited to the area captured by the aerial imagery in Agia Napa and limitations posed to Structure-from-Motion and Multi-view Stereo (SfM-MVS) due to areas of pure texture in Puck Lagoon. Additionally, MagicBathyNet contains 533 annotated patches for seabed habitat and type, facilitating supervised pixel-based classification. Each patch covers an area of 180x180m, represented by 18x18 pixels in S2 imagery, 30x30 pixels in SPOT-6 imagery and 720x720 pixels in airborne imagery. An example of patches is shown in Fig.\ref{fig:fig2} and Fig.\ref{fig:fig3}.

\vspace{-0.2in}
\subsection{Image data collection and processing}
For Agia Napa, imagery from S2 was captured on January 10, 2016, SPOT-6 on January 29 of the same year, and aerial imagery from February 16 to March 16, 2015. In Puck Lagoon, S2 imagery was taken on April 20, 2021, SPOT-6 on April 19 of the same year, and aerial imagery from February 27 to March 2, 2022. The S2 Level2A images were downloaded from Copernicus Hub while the ORTHO SPOT-6 data were downloaded from Airbus Hub via ESA's TPM programme. All the considered satellite images were radiometrically, geometrically, and atmospherically corrected. In Agia Napa, aerial image collection was conducted using a fixed-wing UAV equipped with a Canon IXUS 220HS camera, featuring a 4.3mm focal length, 1.55$\mu$m pixel size, and a 4000×3000 pixels format. A total of 383 images were captured from an average flying height of 209m, resulting in an average ground sampling distance (GSD) of 6.3cm. Georeferencing utilized 40 control points situated solely on land, achieving root mean square errors (RMSEs) of 5.03cm, 4.74cm, and 7.36cm in X, Y, and Z, respectively, with an average reprojection error of 1.11 pixels post-adjustment. In Puck Lagoon, aerial imagery was obtained using a Phase One iXM-100 camera with a 35mm focal length, 3.7$\mu$m pixel size, and a 11664x8750 pixels format. A total of 4384 images were captured from an average flying height of 676m, resulting in an GSD of 6.8cm. Bundle adjustment utilized 22 control points situated solely on land, achieving RMSEs of 1.15cm, 0.71cm, and 0.69cm in X, Y, and Z, respectively, with an average reprojection error of 1.19 pixels post-adjustment. To generate the orthoimages cropped in patches, both SfM-based DSM and aerial imagery were corrected by the refraction effects using the state-of-the-art methods proposed in \cite{agrafiotis2019, agrafiotis2020, agrafiotis2021}. The orthoimages were then divided into non-overlapping patches of 180x180m for each modality. A water mask was created by the near-infrared band of SPOT-6 image in order to get only patches depicting water. Patches in the border of the areas were eliminated.

\vspace{-0.1in}
\subsection{Depth data collection and seabed annotation}

In Agia Napa, the seabed reaches a depth of -30.29m, and bathymetric LiDAR data were collected using the Leica HawkEye III system. In Puck Lagoon, the seabed reaches a depth of -10.57m, and measurements obtained using the Riegl VQ-880-GII LiDAR and Teledyne Reson T50/T20 multibeam echo-sounders \cite{puck, janowski2024}. For seabed annotation, we labeled only pixels that have the same class spanning to all the modalities (S2, SPOT-6 and aerial), considering limitations and differences in spatial resolution. As a consequence, sparse but accurate manual annotations were provided, being the same for each modality and with high confidence level. For each area and modality, around 700 pixels (samples) were manually annotated, belonging to five different classes: i) seagrass (Posidonia oceanica); ii) macroalgae (Filamentous/turf algae); iii) eelgrass/pondweed (Zostera marina, Stuckenia pectinata, Potamogeton perfoliatus); iv) sand; and v) rock. It is noted that the Puck Lagoon area primarily contains classes that are also covered by algal mats, resulting in the prevalent greenish hues seen in most of the annotated pixels. Both depths and seabed annotations were converted into raster format with pixel size equal to the respective modality and then cropped into the non-overlapping 180x180 meters-sized patches. After cropping, each patch was available for additional visual inspection. Table \ref{classes} presents the number of samples of each seabed class and that of annotated image patches per site in MagicBathyNet. The remaining high number of non-annotated patches is also released to enable pretraining with unsupervised learning methods. In the Agia Napa area, the number of patches with annotated samples is smaller due to the limited area covered by the aerial images. However, the seabed's complexity is sufficient to represent all classes and the variations within each class due to the depth.

\vspace{-0.15in}
\begin{table}[h!]
  
  \renewcommand{\arraystretch}{1}
  \small
  \caption{The number of samples per class in MagicBathyNet.}
  \centering
  \label{classes}
  \begin{tabular}{lccc}
    \hline
    \textbf{Class} & \textbf{Agia Napa} & \textbf{Puck Lagoon} & \textbf{Total samples} \\
    \hline
    Seagrass & 118 & 0 & 118  \\
    Macroalgae & 82 & 0 & 82   \\
    Eelgrass etc.& 0 & 428 & 428  \\
    Sand & 246 & 326 & 572  \\
    Rock & 251 & 0 & 251  \\
    \hline
    \# of Samples & 697 & 754 & 1451\\ 
    \# of Patches & 35 & 498 & - \\ 
    \hline
  \end{tabular}
  \vspace{-0.15in}
\end{table}

\vspace{-0.05in}
\section{EXPERIMENTAL RESULTS}
\label{sec:exper}
\vspace{-0.05in}
MagicBathyNet was used for bathymetry retrieval and pixel classification problems. To evaluate the bathymetric performance we considered U-Net\cite{unet} and modified it for estimating continuous values of water depth from RGB images. Excluding the initial and final convolutions, this lighter version comprises of four down-sampling layers with ReLU activation functions, using max pooling for spatial reduction and feature extraction, followed by four up-sampling layers employing transposed convolutions for dimensionality recovery. Additionally, skip connections maintain spatial information during depth prediction. The final convolutional layer in the decoder path has a kernel size of 1x1 and outputs the predicted bathymetry map. Evaluation metrics include RMSE, mean absolute error (MAE) and standard deviation (Std.). For training, RMSE loss that masks out areas with no depths and Adam \cite{kingma2014adam} optimizer were used. The initial learning rate was $10^{-4}$ for SPOT-6 and S2, and $10^{-5}$ for aerial data for a 10-epoch training period. For S2 and aerial data, the learning rate was decreased by a factor of 10 after 9 epochs. 

For pixel-based classification we used U-Net \cite{unet} and SegFormer (B5-sized) \cite{segformer}, an hierarchical Transformer encoder with an all-MLP decode head. We assessed the performance using total accuracy and $F\textsubscript{1}$ score. Stochastic gradient descent (SGD) was the optimizer with a learning rate of $10^{-5}$ for U-Net, and $10^{-4}$ for SegFormer in Agia Napa. In Puck Lagoon, a learning rate of $10^{-5}$ was used for U-Net with SPOT-6 and S2 data, and $10^{-6}$ for aerial. For SegFormer a learning rate of $10^{-4}$ for SPOT-6 and S2, and $10^{-5}$ for aerial was used. All data were trained for 100 epochs. Cross Entropy served as the loss function. For training the models with S2 and SPOT-6 data, for both tasks, resized 256x256 crops were used, notably improving performance without adding new information. The code is implemented in PyTorch, utilizing a single NVIDIA RTX A5000 24 GB GPU for training and testing. The dataset was divided into random splits (80\% training, 20\% evaluation), customized for pixel-based classification and bathymetry prediction, ensuring reproducibility.

Table \ref{archive_list} shows results attained by modified U-Net on the MagicBathyNet dataset for predicting bathymetry in the two different areas, across the three modalities. It is evident that aerial data shows the best predictive performance, with lower RMSE, MAE, and Std. values in both Puck Lagoon and Agia Napa compared to SPOT-6 and S2. In Puck Lagoon, the model achieves better performance metrics for aerial and S2 modalities compared to Agia Napa. The significantly higher RMSE in Agia Napa's aerial data, compared to Puck Lagoon, is due to the sparse depth labels of the aerial imagery specifically in this area.  For SPOT-6, the performance is slightly better in Agia Napa in terms of RMSE and Std. but is relatively comparable between the two sites. Results indicate that Agia Napa is a more challenging site due to variable seabed cover, abrupt depth changes, and greater depth. 

\vspace{-0.15in}
\begin{table}[h!]
  \setlength{\tabcolsep}{1.5pt}
  \renewcommand{\arraystretch}{1}
  \small
  \caption{Results obtained by modified U-Net (in meters).}
  \centering
  \label{archive_list}
  \begin{tabular}{m{20mm}|>{\centering\arraybackslash}m{10mm}>{\centering\arraybackslash}m{11mm}>{\centering\arraybackslash}m{7mm}|>{\centering\arraybackslash}m{10mm}>{\centering\arraybackslash}m{11mm}>{\centering\arraybackslash}m{7mm}}
  \hline 

  \textbf{Modality}                   & \multicolumn{3}{c|}{\textbf{Agia Napa}} & \multicolumn{3}{c}{\textbf{Puck Lagoon}} \\
                     \cline{2-7}
                     & \textbf{RMSE} & \textbf{MAE}& \textbf{Std.} & \textbf{RMSE} & \textbf{MAE}& \textbf{Std.}\\
                     \hline
  aerial&0.616&0.420&0.598&0.298&0.187&0.297 \\
  \hline
  SPOT-6 &0.718&0.483&0.691&0.817&0.412&0.815 \\
  \hline
  S2 &1.068&0.694&0.940&0.907&0.493&0.874 \\
  \hline 
  \end{tabular}
\vspace{-0.1in}
\end{table}

Tables \ref{archive_list2}, \ref{archive_list3}, and \ref{archive_list4} as well as Fig.\ref{fig:fig2} and Fig.\ref{fig:fig3}, show the pixel-based classification results. Both models show a similar pattern and varied performance across classes, areas, and modalities. While some classes achieve consistently high $F\textsubscript{1}$ Scores (i.e. seagrass, eelgrass/pondweed and sand), others show variability in performance based on the specific modality and area (i.e. macroalgae), influencing the total accuracy. 

\begin{table}[h!]
\vspace{-0.15in}
  \setlength{\tabcolsep}{1.5pt}
  \renewcommand{\arraystretch}{1}
  \small
  \caption{$F\textsubscript{1}$ scores (\%) obtained in Agia Napa.}
  \centering
  \label{archive_list2}
  \begin{tabular}{m{20mm}|>{\centering\arraybackslash}m{10mm}>{\centering\arraybackslash}m{11mm}>{\centering\arraybackslash}m{7mm}|>{\centering\arraybackslash}m{10mm}>{\centering\arraybackslash}m{11mm}>{\centering\arraybackslash}m{7mm}}
  \hline 

  \textbf{Classes}                   & \multicolumn{3}{c|}{\textbf{U-Net}} & \multicolumn{3}{c}{\textbf{SegFormer}} \\
                     \cline{2-7}
                     & \textbf{aerial} & \textbf{SPOT-6}& \textbf{S2} & \textbf{aerial} & \textbf{SPOT-6} & \textbf{S2}\\
                     \hline
   
  \hline
  Seagrass &97.37&83.33&84.21&97.37&77.92&74.70 \\
  \hline
  Macroalgae &94.73&13.33&25.00&87.50&46.15&50.00 \\
  \hline
  Sand &93.33&85.11&90.70&93.48&80.43&76.31 \\
  \hline
  Rock &92.63&70.80&74.78&94.83&77.97&78.99 \\
  \hline 
  \end{tabular}
  \vspace{-0.15in}
\end{table}

\begin{table}[h!]
  \vspace{-0.15in}
  \setlength{\tabcolsep}{1.5pt}
  \renewcommand{\arraystretch}{1}
  \small
  \caption{$F\textsubscript{1}$ scores (\%) obtained in Puck Lagoon.}
  \centering
  \label{archive_list3}
  \begin{tabular}{m{20mm}|>{\centering\arraybackslash}m{10mm}>{\centering\arraybackslash}m{11mm}>{\centering\arraybackslash}m{7mm}|>{\centering\arraybackslash}m{10mm}>{\centering\arraybackslash}m{11mm}>{\centering\arraybackslash}m{7mm}}
  \hline 

  \textbf{Classes}                   & \multicolumn{3}{c|}{\textbf{U-Net}} & \multicolumn{3}{c}{\textbf{SegFormer}} \\
                     \cline{2-7}
                     & \textbf{aerial} & \textbf{SPOT-6}& \textbf{S2} & \textbf{aerial} & \textbf{SPOT-6} & \textbf{S2}\\
                     \hline
  Eelgrass etc.&95.36&87.91&79.75&97.47&89.01&83.54 \\
  \hline
  Sand &95.30&81.36&77.14&97.18&83.05&81.43 \\
  \hline
  \end{tabular}
  \vspace{-0.15in}
\end{table}

\begin{table}[h!]
\vspace{-0.15in}
  \setlength{\tabcolsep}{1.5pt}
  \renewcommand{\arraystretch}{1}
  \small
  \caption{Total Accuracy (\%) obtained in both sites.}
  \centering
  \label{archive_list4}
  \begin{tabular}{m{20mm}|>{\centering\arraybackslash}m{10mm}>{\centering\arraybackslash}m{11mm}>{\centering\arraybackslash}m{7mm}|>{\centering\arraybackslash}m{10mm}>{\centering\arraybackslash}m{11mm}>{\centering\arraybackslash}m{7mm}}
  \hline 

  \textbf{Area}                   & \multicolumn{3}{c|}{\textbf{U-Net}} & \multicolumn{3}{c}{\textbf{SegFormer}} \\
                     \cline{2-7}
                     & \textbf{aerial} & \textbf{SPOT-6}& \textbf{S2} & \textbf{aerial} & \textbf{SPOT-6} & \textbf{S2}\\
                     \hline
  Agia Napa&94.67&76.00&78.76&94.67&77.33&75.51 \\
  \hline
  Puck Lagoon &95.33&85.33&78.52&97.33&86.66&82.55 \\
  \hline
  \end{tabular}
\vspace{-0.1in}
\end{table}

SegFormer outperformed U-Net in $F\textsubscript{1}$ scores, except for seagrass and sand classes in SPOT-6 and S2 data, and the macroalgae class in aerial data, all in Agia Napa area. The reduced performance of SegFormer can be attributed to the reduced number of training samples. In contrast, U-Net is better suited for situations with limited data, as it effectively leverages features at various levels of abstraction. For seagrass classification in aerial data from Agia Napa, both networks performed similarly. The varied performance observed in classifying macroalgae in SPOT-6 and S2 data can be attributed to the limited coverage area of the class on the seabed. In most cases, this area is considerably smaller than the GSD of both satellite images. However, this is not the case in the aerial data. Visual inspection of the predicted seabed classes was also performed, revealing differences in predictions across non-annotated areas. As shown in Fig. \ref{fig:fig2}, U-Net classified pixels as macroalgae and sand where only rock exists. Similarly, in Fig. \ref{fig:fig3} is shown that U-Net classified sandy areas as eelgrass/pondweed. These differences highlight SegFormer's superior accuracy, indicated by higher $F\textsubscript{1}$ scores, attributed to its architecture which captures long-range dependencies and global context better than traditional architectures like U-Net. This leads to more accurate predictions, especially in unlabeled regions. Utilizing deep models trained on the unlabeled patches of MagicBathyNet can yield more favorable outcomes, particularly noticeable in SPOT-6 and S2.

\begin{figure}[t!]
  \setlength{\tabcolsep}{1.5pt}
  \renewcommand{\arraystretch}{1}
  \footnotesize
  
  \centering

  \begin{tabular}{m{0.5cm}ccc}
 
   \textbf{} & \textbf{aerial} & \textbf{SPOT-6} & \textbf{S2} \\

   \textbf{(a)} &\begin{minipage}[c]{0.27\columnwidth}
        \centering
        \includegraphics[width=\linewidth]{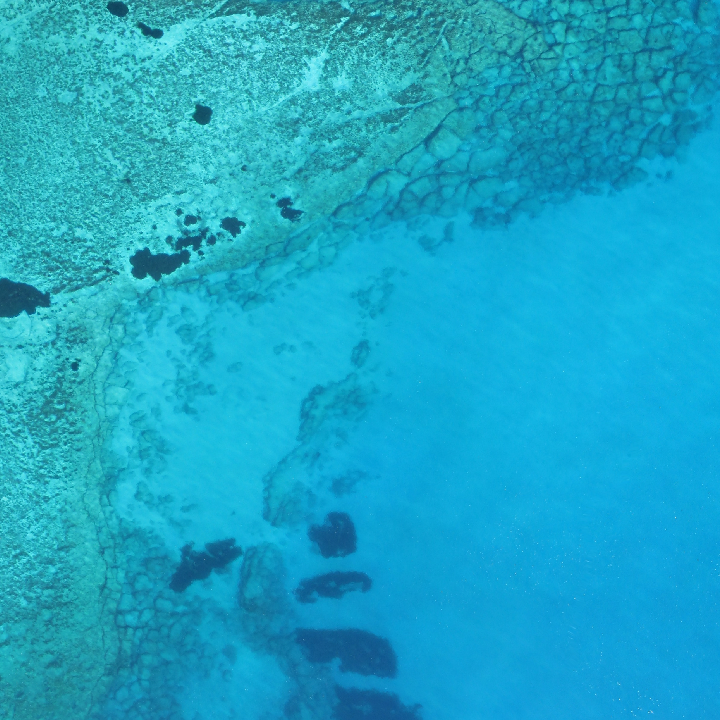}
    \end{minipage}& 
    \hfill

    \begin{minipage}[c]{0.27\columnwidth}
        \centering
        \includegraphics[width=\linewidth]{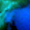}
    \end{minipage}&    
    \hfill

    \begin{minipage}[c]{0.27\columnwidth}
        \centering
        \includegraphics[width=\linewidth]{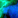}
    \end{minipage} \vspace{1pt}\\

     \textbf{(b)} &\begin{minipage}[c]{0.27\columnwidth}
        \centering
        \includegraphics[width=\linewidth]{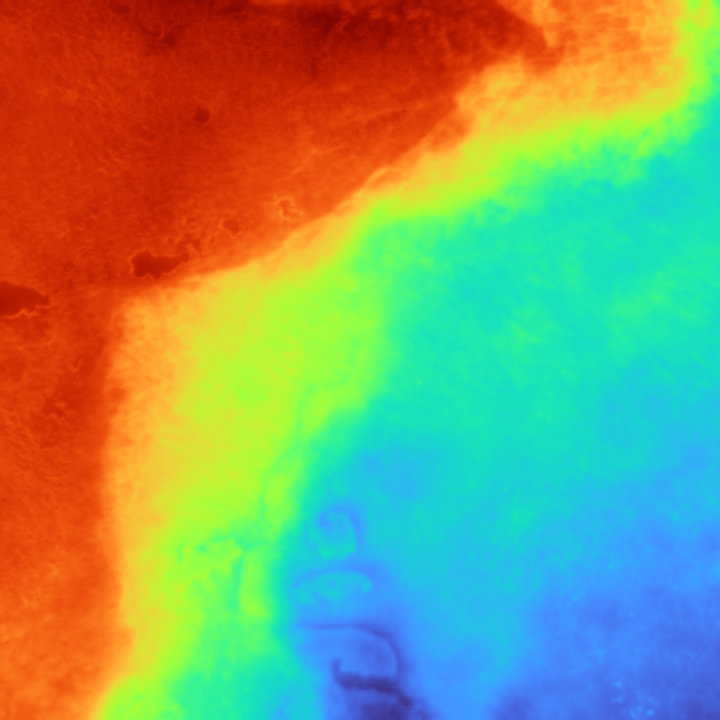}
    \end{minipage}& 
    \hfill

    \begin{minipage}[c]{0.27\columnwidth}
        \centering
        \includegraphics[width=\linewidth]{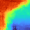}
    \end{minipage}&    
    \hfill

    \begin{minipage}[c]{0.27\columnwidth}
        \centering
        \includegraphics[width=\linewidth]{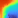}
    \end{minipage} \vspace{1pt}\\

  \textbf{(c)} &\begin{minipage}[c]{0.27\columnwidth}
        \centering
        \includegraphics[width=\linewidth]{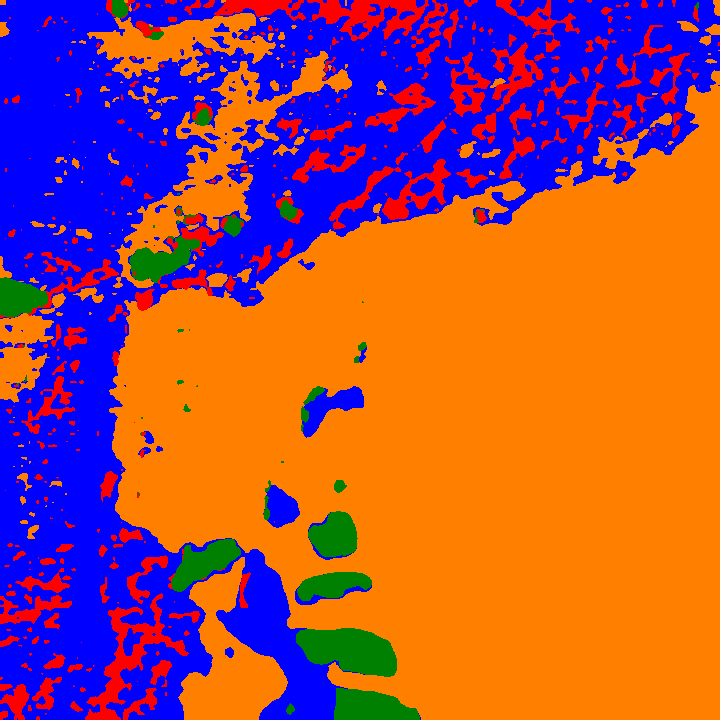}
    \end{minipage}& 
    \hfill

    \begin{minipage}[c]{0.27\columnwidth}
        \centering
        \includegraphics[width=\linewidth]{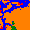}
    \end{minipage}&    
    \hfill

    \begin{minipage}[c]{0.27\columnwidth}
        \centering
        \includegraphics[width=\linewidth]{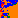}
    \end{minipage} \vspace{1pt}\\

  \textbf{(d)} &\begin{minipage}[c]{0.27\columnwidth}
        \centering
        \includegraphics[width=\linewidth]{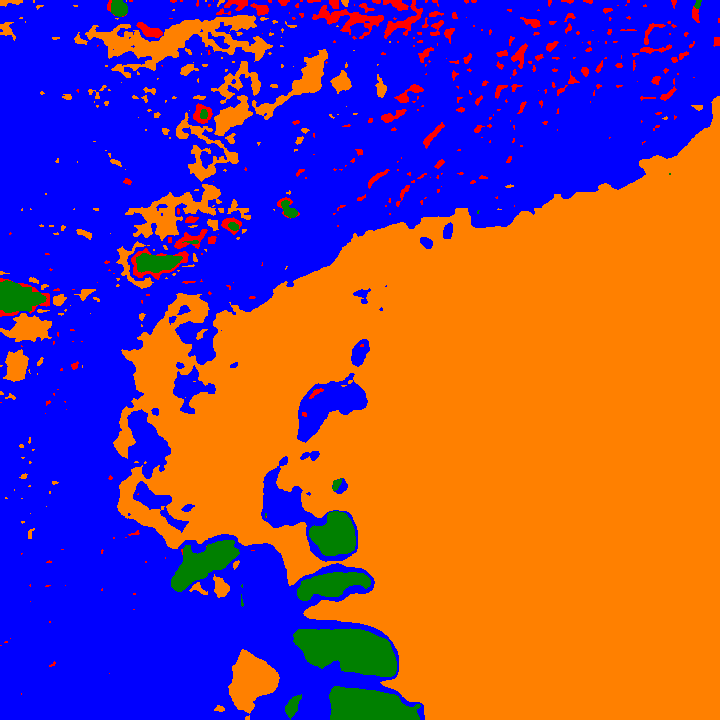}
    \end{minipage}& 
    \hfill

    \begin{minipage}[c]{0.27\columnwidth}
        \centering
        \includegraphics[width=\linewidth]{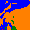}
    \end{minipage}&    
    \hfill

    \begin{minipage}[c]{0.27\columnwidth}
        \centering
        \includegraphics[width=\linewidth]{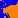}
    \end{minipage} \vspace{1pt}\\

    \textbf{} &
    \multicolumn{3}{l}
    
    \begin{minipage}[c]{0.77\columnwidth}
        \centering
        \includegraphics[width=\linewidth]{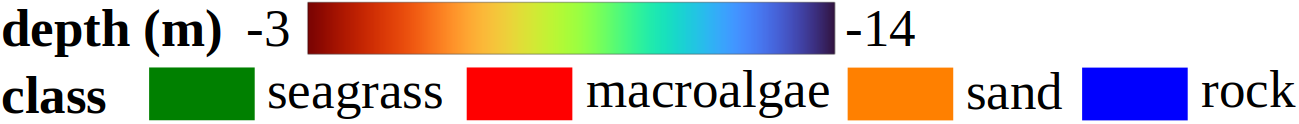}
\end{minipage}\\
\end{tabular}

\caption{(a) True color composite of example patches acquired over Agia Napa, (b) bathymetry obtained by modified U-Net, seabed classes obtained by (c) U-Net and (d) SegFormer.}
  \label{fig:fig2}
  \vspace{-0.15in}
\end{figure}

\begin{figure}[h!]
  \setlength{\tabcolsep}{1.5pt}
  \renewcommand{\arraystretch}{1}
  \footnotesize
  \centering

  \begin{tabular}{m{0.5cm}ccc}
      \textbf{} & \textbf{aerial} & \textbf{SPOT-6} & \textbf{S2} \\
     \textbf{(a)}
    &\begin{minipage}[c]{0.27\columnwidth}
        \centering
        \includegraphics[width=\linewidth]{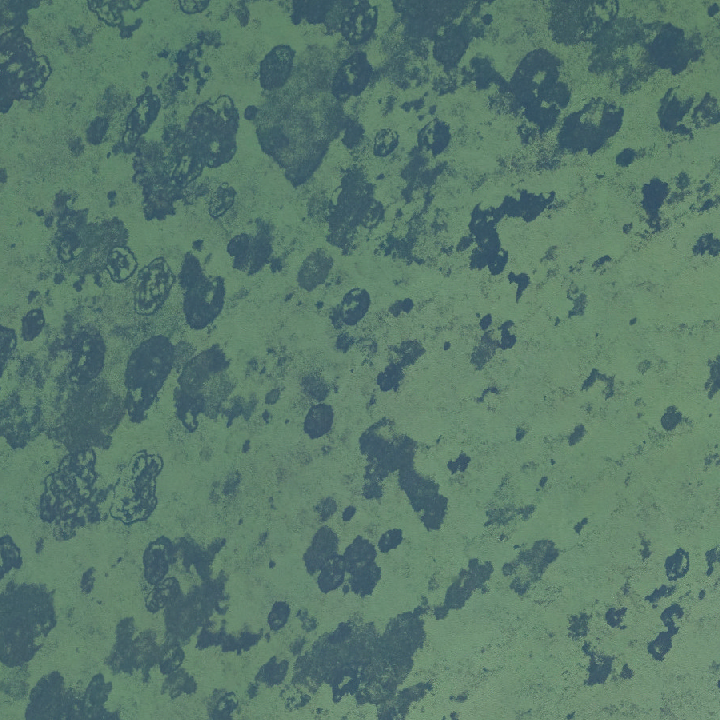}
    \end{minipage}& 
    \hfill

    \begin{minipage}[c]{0.27\columnwidth}
        \centering
        \includegraphics[width=\linewidth]{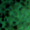}
    \end{minipage}&    
    \hfill

    \begin{minipage}[c]{0.27\columnwidth}
        \centering
        \includegraphics[width=\linewidth]{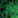}
    \end{minipage} \vspace{1pt}\\
     \textbf{(b)}  &\begin{minipage}[c]{0.27\columnwidth}
        \centering
        \includegraphics[width=\linewidth]{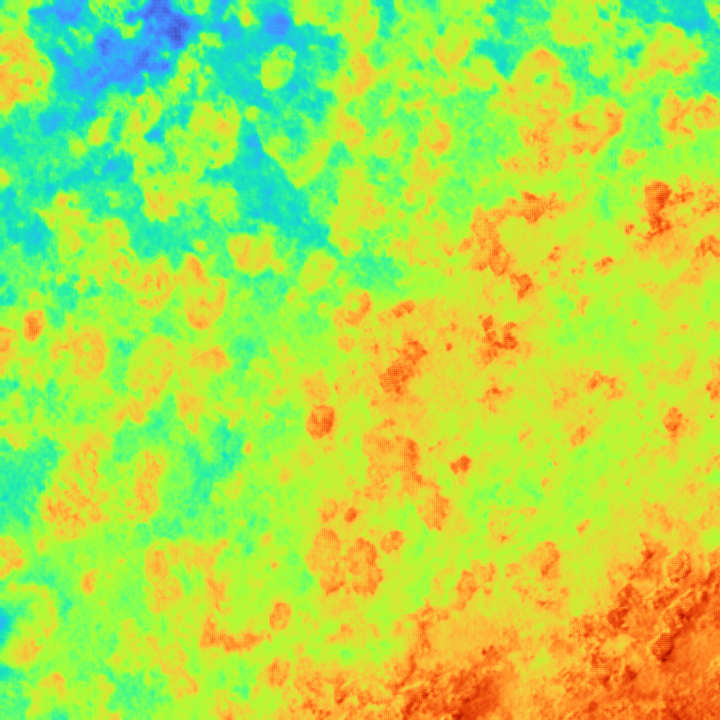}
    \end{minipage}& 
    \hfill

    \begin{minipage}[c]{0.27\columnwidth}
        \centering
        \includegraphics[width=\linewidth]{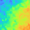}
    \end{minipage}&    
    \hfill

    \begin{minipage}[c]{0.27\columnwidth}
        \centering
        \includegraphics[width=\linewidth]{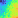}
    \end{minipage} \vspace{1pt}\\
  \textbf{(c)}   &\begin{minipage}[c]{0.27\columnwidth}
        \centering
        \includegraphics[width=\linewidth]{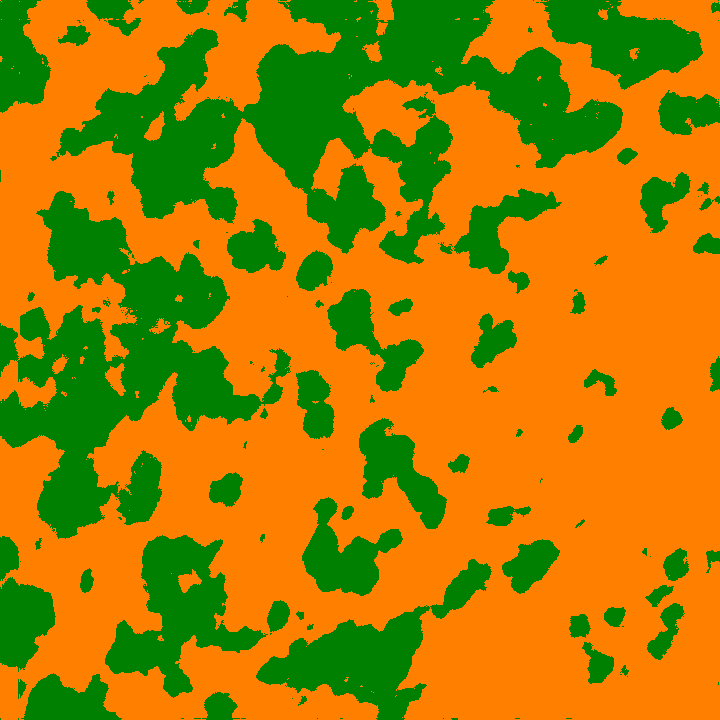}
    \end{minipage}& 
    \hfill

    \begin{minipage}[c]{0.27\columnwidth}
        \centering
        \includegraphics[width=\linewidth]{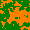}
    \end{minipage}&    
    \hfill

    \begin{minipage}[c]{0.27\columnwidth}
        \centering
        \includegraphics[width=\linewidth]{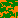}
    \end{minipage} \vspace{1pt}\\
    \textbf{(d)} &\begin{minipage}[c]{0.27\columnwidth}
        \centering
        \includegraphics[width=\linewidth]{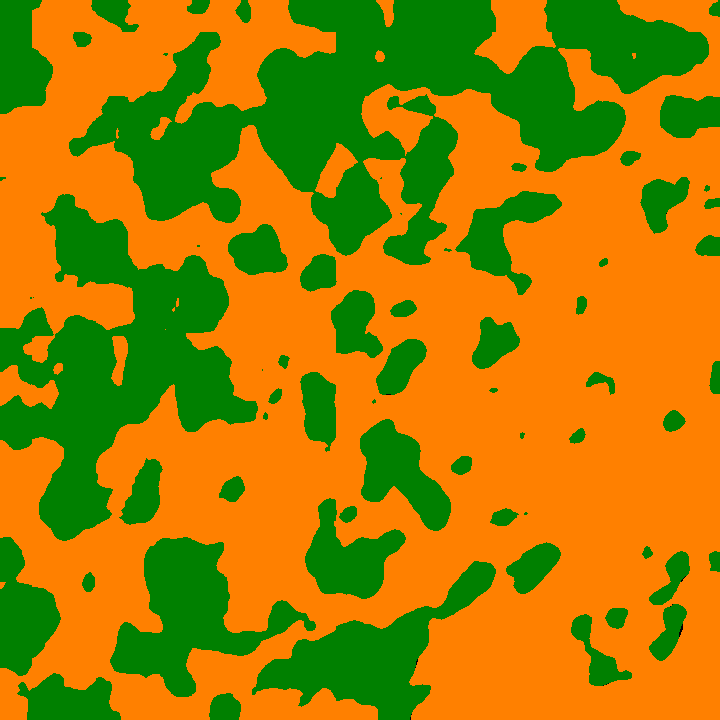}
    \end{minipage}& 
    \hfill

    \begin{minipage}[c]{0.27\columnwidth}
        \centering
        \includegraphics[width=\linewidth]{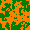}
    \end{minipage}&    
    \hfill

    \begin{minipage}[c]{0.27\columnwidth}
        \centering
        \includegraphics[width=\linewidth]{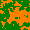}
    \end{minipage} \vspace{1pt}\\
   \textbf{} &
    \multicolumn{3}{l}
    
    \begin{minipage}[c]{0.77\columnwidth}
        \centering
        \includegraphics[width=\linewidth]{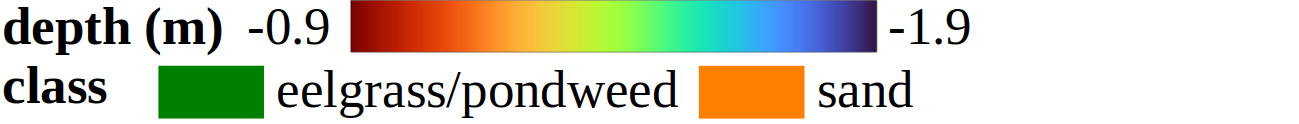}
    \end{minipage} \\
  \end{tabular}

\caption{(a) True color composite of example patches acquired over Puck Lagoon, (b) bathymetry obtained by modified U-Net, seabed classes obtained by (c) U-Net and (d) SegFormer.}
\vspace{-0.15in}
  \label{fig:fig3}
\end{figure}

\vspace{-0.05in}
\section{CONCLUSION}
\label{sec:con}
\vspace{-0.05in}

This paper has introduced MagicBathyNet, which is a benchmark dataset made up of image patches of S2, SPOT-6, aerial imagery, bathymetry and seabed annotations for learning-based bathymetric mapping and pixel-based classification. To the best of our knowledge, MagicBathyNet is the first publicly available multimodal dataset for these tasks, aiming to enable substantial progress in seabed mapping with deep learning. MagicBathyNet also includes a high number of unlabeled samples in addition to the labeled data. These unlabeled samples can be used for pre-training through self-supervision, followed by fine-tuning on labeled data for various downstream tasks. As part of future work, exploring multimodal learning techniques is anticipated to further enhance the dataset's utility. Additionally, we plan to expand the dataset to cover more geographical areas, representing a wider range of water and seabed types.

\vspace{+0.04in}
\section{ACKNOWLEDGMENTS}
\label{sec:ackn}

This work is part of MagicBathy project funded by the European Union’s HORIZON Europe research and innovation programme under the Marie Skłodowska-Curie Actions agreement No 101063294. The authors thank the European Space Agency (TPM programme PP0092443) and Airbus for SPOT-6 imagery, NVIDIA Corporation for support via NVIDIA Academic Hardware Grant Program, and the Department of Land and Surveys of Cyprus for providing the LiDAR data of Agia Napa. L. Janowski was funded by the NSC, Poland (GN 2021/40/C/ST10/00240). Dr. Poursanidis is acknowledged for identifying filamentous/turf algae.


\newpage
\bibliographystyle{IEEEbib}
\bibliography{refs}
\vfill

\end{document}